\documentclass[runningheads]{llncs}
\usepackage[T1]{fontenc}
\usepackage{amsmath, array}
\usepackage{graphicx,verbatim}
\usepackage{color}
\usepackage{caption}
\usepackage{float}
\usepackage{arydshln}
\usepackage[table]{xcolor}
\usepackage{amsfonts}
\usepackage{algorithm}
\usepackage[noend]{algpseudocode}
\usepackage{hyperref}
\usepackage[htt]{hyphenat}
\usepackage{color}
\usepackage{booktabs}
\usepackage{multirow}

\newcommand{\ourmethod}{\textit{AURA}}

\begin{document}
\title{AURA: A Multi-Modal Medical Agent for Understanding, Reasoning \& Annotation}
\titlerunning{AURA: Agent for Understanding, Reasoning \& Annotation}

\author{Nima Fathi\inst{1, 2} \and
Amar Kumar\inst{1, 2} \and
Tal Arbel\inst{1,2}}
\authorrunning{N. Fathi et al.}
%
\institute{Center for Intelligent Machines, McGill University, Montreal, Canada \and
Mila - Quebec AI institute, Montreal, Canada \\
\email{nima.fathi@mila.quebec}}

\maketitle              
\begin{abstract}

Recent advancements in Large Language Models (LLMs) have catalyzed a paradigm shift from static prediction systems to agentic AI—intelligent agents capable of reasoning, interacting with tools, and adapting to complex tasks. While LLM agentic systems have shown promise across many domains, their application to medical imaging remains in its infancy. In this work, we introduce \ourmethod, the first visual linguistic explainability agent designed specifically for comprehensive analysis, explanation, and evaluation of medical images. By enabling dynamic interactions, contextual explanations, and hypothesis testing, \ourmethod\ represents a significant advancement toward more transparent, adaptable, and clinically aligned AI systems. This work highlights the promise of agentic AI in transforming medical image analysis from static predictions to interactive decision support. Leveraging Qwen-32B, an LLM-based architecture, \ourmethod\ integrates a modular toolbox comprising: (i) a segmentation suite with phase grounding, pathology segmentation, and anatomy segmentation to localize clinically meaningful regions; (ii) a counterfactual image generation module that supports reasoning through image-level explanations; and (iii) a set of evaluation tools, including pixel-wise difference map analysis, classification, and advanced state-of-the-art components, to assess the diagnostic relevance and visual interpretability of the results. Our code is accessible through
the project website.~\footnote{\href{https://nimafathi.github.io/AURA/}{https://nimafathi.github.io/AURA/}}

\end{abstract}


\section{Introduction}
Conventional AI models in medical imaging often fail to meet the needs of real clinical practice. Ideally, an AI system would reason independently, recognize when it lacks sufficient context, and dynamically use various tools—much like clinicians do in complex diagnostic scenarios. However, traditional medical imaging AI is typically rigid, designed for specific tasks with fixed inputs and outputs. This lack of flexibility prevents these systems from adapting to changing clinical situations. When faced with unclear findings, unfamiliar diseases, or incomplete information, these models cannot ask for additional details, gather more data, or revise their conclusions~\cite{sapkota2025ai,shavit2023practices}. Consequently, they fall short in interpretability, adaptability, and gaining clinical trust. Agentic AI offers a promising alternative, providing models that not only handle specific tasks but also reason through uncertainty, generate clear visual and linguistic explanations (VLEs), test hypotheses via counterfactual simulations, and collaborate interactively with clinicians. By combining autonomous reasoning with dynamic tool use, agentic systems significantly enhance AI's practical utility in clinical settings, bridging the gap between static automation and the flexible decision-making required in medical practice.

Agentic AI has gained significant popularity across a variety of domains where systems must reason under uncertainty, take autonomous actions, and interact adaptively with their environments~\cite{acharya2025agentic,hughes2025ai}. Web-based agents such as AutoGPT~\cite{yang2023auto} and Voyager~\cite{wang2023voyager} have demonstrated how large language models (LLMs) can control digital interfaces, pursue multi-step goals, and coordinate tool use—shifting the role of AI from static prediction to dynamic, goal-driven problem-solving. In medical imaging, the principles of agentic AI are beginning to shape how multiple models and tools can work together coherently to assist with complex clinical tasks~\cite{zou2025rise,karunanayake2025next}. Foundation models (FMs), including LLMs and large multimodal models (LMMs), provide a powerful foundation for building unified frameworks that integrate medical image-text reasoning, diagnostic analysis, and clinical decision support. Several recent agentic frameworks have advanced this direction by introducing structured reasoning, tool orchestration, and modality-aware workflows. MDAgents\cite{kim2024mdagents} presents a multi-agent system that dynamically configures collaboration between LLMs based on task complexity, achieving state-of-the-art (SOTA) performance on medical benchmarks. MMedAgent\cite{li2024mmedagent} introduces the first multi-modal medical AI agent capable of intelligently selecting and integrating specialized tools—such as segmentation, classification, and report generation—across five imaging modalities. Using instruction tuning and dynamic tool invocation, it outperforms both open- and closed-source models, including GPT-4o~\cite{hurst2024gpt}. MedRAX~\cite{fallahpour2025medrax} combines state-of-the-art analysis tools with multimodal LLMs to address complex queries involving visual question answering and report generation. While these frameworks represent major progress, many still lack explicit reasoning capabilities and robust image-grounded explanations—features critical for transparency, safety, and clinical trust. Together, these frameworks underscore the increasing role of agentic AI in developing more interactive and intelligent healthcare systems. 

In this work, we present \ourmethod, the first AI agent designed to analyze, generate visual-linguistic explanations (VLEs), and perform self-evaluations using a comprehensive suite of SOTA tools. Unlike conventional models that offer limited interpretability and operate in a static inference paradigm, \ourmethod\ emphasizes dynamic VLE, introspective evaluation, and adaptive reasoning in data-scarce clinical settings. Our contributions include an agent capable of: (i) image segmentation with phase grounding (associating regions of medical image to corresponding text or clinical concepts), pathology segmentation, and anatomy segmentation to identify and attribute clinically relevant regions; (ii) CF image generation for probing understanding through controlled perturbations; and (iii) a self-evaluation toolkit incorporating difference map analysis, classification, and specialized tools on Chest X-ray (CXR) images such as RadEdit~\cite{perez2025radedit} and PRISM~\cite{kumar2025prism}, enabling critical assessment of its own outputs. Notably, \ourmethod\ performs robustly even in low-supervision scenarios (less textual prompt guidance) by autonomously identifying knowledge gaps, invoking report generation to gain context, and generating multiple candidate CFs. Finally, the best candidate is chosen using a self-evaluation scoring mechanism. This integration of explanation, self-assessment, and context-aware tool use marks a step toward more trustworthy, adaptable, and clinically actionable AI systems.

\section{AURA: An Agent for Visual-Linguistic Explanation}
\label{sec:AURA}

\begin{figure}[!b]
\centering
\includegraphics[scale=0.1]{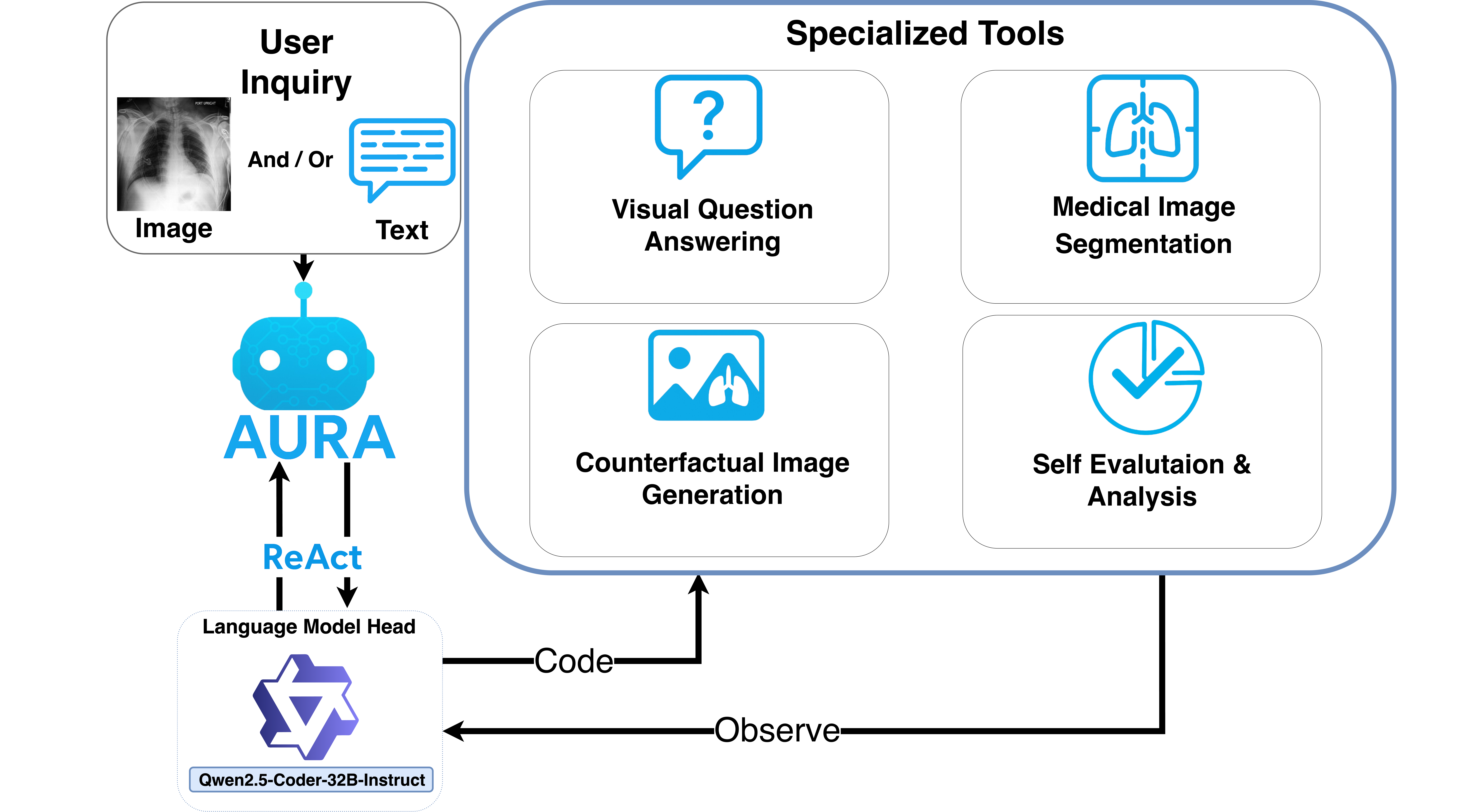}
\caption{\textbf{\ourmethod\ architecture} integrates a reasoning loop that dynamically selects from a multimodal ecosystem of tools. The agent iteratively reasons over image-query pairs, generates \textit{code} to invoke tools, and \textit{observes} their resulting outputs to synthesize high-quality visual and textual medical explanations. ReAct framework enables visualization and control of the agent’s reasoning process.}
\label{fig:aura-architecture}
\end{figure}

We present \ourmethod, an open-source, modular agent designed to provide interpretable, multimodal visual-linguistic explanations for medical imaging. Built to address the limitations of static and non-integrated AI pipelines, \ourmethod\ flexibly integrates a diverse suite of expert medical tools—spanning report generation and grounding, visual question answering (VQA), segmentation, pathology detection, and counterfactual image editing—within a reasoning-driven loop powered by a large language model. 

\ourmethod\ supports user-driven interactions and autonomous self-evaluation, allowing it to select and optimize among multiple strategies to produce clinically meaningful outputs, Figure~\ref{fig:aura-architecture}. We conduct all evaluations and demonstrate the agent’s capabilities in the context of chest X-rays, leveraging the CheXpert dataset~\cite{irvin2019chexpert}.

\begin{algorithm}[H]
    \caption{Overview of \ourmethod's Dynamic Reasoning and Tool Orchestration}
    \label{alg:aura_reasoning_loop}
\begin{algorithmic}[1]
    \State \textbf{Input:} $Q$: User query, $I$: Chest X-ray image (optional), $\mathcal{T}$: Set of available medical AI tools, $M$: LLM Head, $t_{max}$: Maximum allowed steps
    \State \textbf{Initialize:} $step \leftarrow 0$; $memory \leftarrow \emptyset$ ; $current\_state \leftarrow \Call{Observe}{Q, I, memory}$ 

    \While{$step < t_{max}$}
        \State \Comment{Agent reasons about the next action based on current state and memory}
        \State $thought \leftarrow \Call{M.Reason}{current\_state, memory}$

        \If{\Call{M.CanGenerateFinalAnswer}{thought, memory}} \Comment{LLM determines if a final answer can be formed based on memory and the last ``thought"}
            \State \Return \Call{M.GenerateFinalAnswer}{thought, memory}
        \EndIf

        \State \Comment{Select the most appropriate tool based on reasoning and current state}
        \State $tool \leftarrow \Call{SelectTool}{thought, \mathcal{T}, current\_state}$
        \State $tool\_input \leftarrow \Call{PrepareToolInput}{thought, current\_state, I}$ \Comment{Prepare input for the selected tool}

        \State \Comment{Execute the selected tool}
        \State $tool\_result \leftarrow \Call{Execute}{tool, tool\_input}$

        \State \Comment{Update memory and current state with the result of tool execution}
        \State $memory \leftarrow memory \cup \{(thought, tool, tool\_result)\}$
        \State $current\_state \leftarrow \Call{M.Observe}{current\_state, tool\_result, memory}$
        \State $step \leftarrow step + 1$
    \EndWhile

    \State \Return \Call{M.GenerateTimeoutResponse}{current\_state, memory}
\end{algorithmic}
\end{algorithm}

\subsection{Dynamic Reasoning with an Expert Toolkit}
\ourmethod\ is driven by a ReAct-style reasoning loop~\cite{yao2023react} powered by a code-instructed LLM, which is specifically fine-tuned and optimized for understanding and generating executable code. This enables the agent to break down a user's request into a sequence of logical steps and execute and perform each step by leveraging its integrated set of medical tools. (Algorithm \autoref{alg:aura_reasoning_loop}). Unlike static pipelines, our dynamic approach enables \ourmethod\ to infer chest CXR findings—crucial for visual verification and clinical interpretability—and to dynamically generate and select the most relevant visual evidence through a self-evaluation mechanism that effectively supports its textual responses. \ourmethod's analytical power comes from its ability to flexibly chain tools from its integrated toolkit:
\begin{itemize}
    \item \textbf{Visual Question Answering}: Radiology specific dialogue using the ChexAgent VQA~\cite{chen2024chexagent} or generates pathology medical reports using MAIRA-2~\cite{bannur2024maira}.
    \item \textbf{Grounded Report Generation}: Employs MAIRA-2~\cite{bannur2024maira} to align medical findings with bounding box or segmentation overlays for visual grounding.
    \item \textbf{Counterfactual Editing}: Utilizes \textit{RadEdit}~\cite{perez2025radedit} for precision-guided image editing of pathologies and \textit{PRISM}~\cite{kumar2025prism} to generate counterfactual images.
    \item \textbf{Segmentation and Detection}: Leverages MedSAM~\cite{ma2024segment} and PSPNet~\cite{pspnet} for anatomy localization and TorchXRayVision for pathology classification~\cite{cohen2022torchxrayvision}.
    \item \textbf{Analysis and Visualization}: Difference maps to quantify edits, generate subject-specific image variations, and manage the overall analysis session.
\end{itemize}

\subsection{Modular Architecture for Self-Evaluation}
\ourmethod's architecture is implemented using modular segments where each tool is an independent component. This design supports parallel execution across multiple GPUs, robust fallback strategies, and makes the system easily extendable. More importantly, this modularity is the foundation for \ourmethod's most defining feature: self-evaluation. By treating its own tools as both actors and critics, \ourmethod\ can autonomously assess and refine its work.
For example, when tasked with removing a pathological structure from an image, the agent initiates the following multi-step, self-correcting workflow:
\begin{enumerate}
    \item Generate several candidate CF images (offering varying explainability) using \textit{RadEdit}~\cite{perez2025radedit} and \textit{PRISM}~\cite{kumar2025prism} with varying hyper-parameters.
    \item Leverage \textit{TorchXRayVision}~\cite{cohen2022torchxrayvision} to classify pathological structures in the original and for each CF, candidate explanation image.
    \item Compare pathology and similarity scores, selecting CF that achieves the highest improvement in the target pathology's score while effectively preserving subject identity, ensuring a clinically relevant and high-quality edit.
\end{enumerate}

\begin{figure}[t!]
\centering
\includegraphics[scale=0.26]{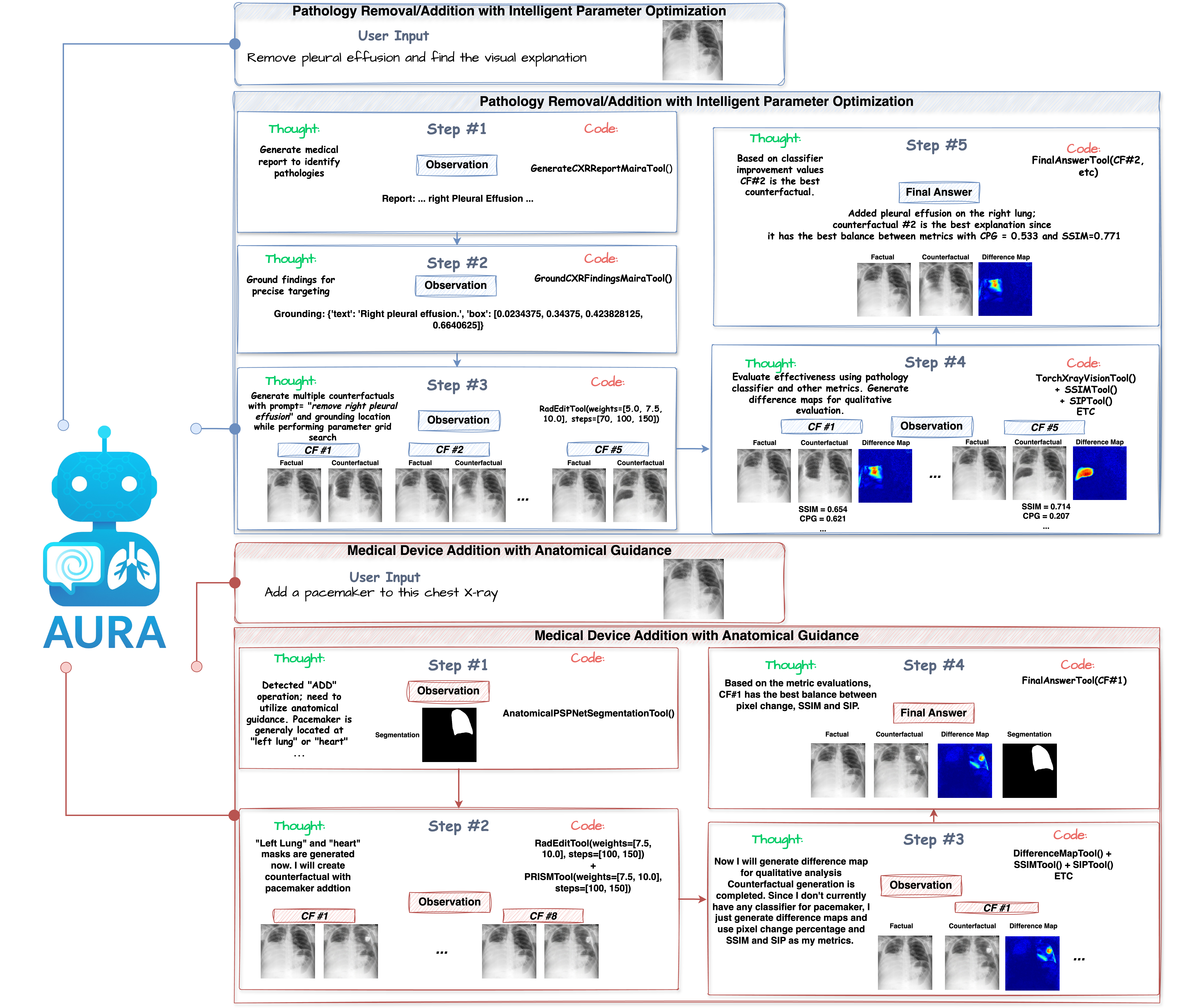}
\caption{\textbf{AURA Explainability Flow.} \ourmethod\ handles an open-ended clinical question by analyzing the image, localizing findings, and producing both visual and linguistic explanations. The interface displays intermediate results and self-evaluations to ensure transparency. Note: The \textcolor[HTML]{00CC66}{thought} behind evaluation and selecting the final image is described in the image.  Zoom in for better visibility.}
\label{fig:aura-interface}
\end{figure}

This generate–test–select strategy is a core capability of \ourmethod, enabling it to optimize parameters, compare outputs across models, and produce accurate, visually coherent explanations. The entire process is made transparent to the user through an interactive interface that displays step-by-step diagnostic evidence, including visual comparisons of counterfactual and factual images. This interface provides both qualitative explanations through difference maps and quantitative justification through corresponding pathology scores. For an overview of different scenarios that our model could handle, please refer to Figure~\ref{fig:aura-interface}.

\section{Experiments and Results}
\subsection{Dataset \& Implementation Details}
For evaluation, we utilize the held-out test set of the publicly available CheXpert dataset~\cite{irvin2019chexpert}. As \ourmethod\ operates as an inference agent, leveraging off-the-shelf tools rather than being fine-tuned, this held-out set provides the sole relevant data for assessing its performance. To ensure a fair comparison and mitigate potential biases or distribution shifts, we adopt the identical data split used by PRISM \cite{kumar2025prism}. \ourmethod\ deploys Qwen2.5-Coder-32B-Instruct~\cite{hui2024qwen2} as its backbone LLM. This leverages Qwen's capabilities in generating executable code actions and robust function calls for seamless tool interaction \cite{wang2024executable}. The agent orchestrates its integrated tools via the SmolAgents framework, allowing it to formulate requests with required arguments through programmatic Python function calls~\cite{smolagents}. \ourmethod \ uses two NVIDIA A100 80GB GPUs as the backend for running inferences. In case multiple GPUs are detected on the system, our method is adaptive to parallelize on multiple GPUs. This on-premises deployment is particularly advantageous in medical settings, as it provides a secure and controlled environment for sensitive patient data and effectively mitigates the privacy risks associated with external, cloud-based LLM APIs.

\subsection{Agentic Image-based Explainability via CF Image Generation}
We demonstrate the efficacy of \ourmethod's capabilities in CF image generation through a comparative experiment against its underlying image editing tools, RadEdit and PRISM. This setup highlights how \ourmethod's autonomous decision-making outperforms fixed-strategy baselines. For RadEdit and PRISM, we evaluate both single CF generation (one CF per instance) and an "Ensemble" approach, where five CFs are generated per instance by varying parameters such as guidance scale and inference steps (with RadEdit also including grounded and ungrounded conditions, where applicable). A post-processing script is then used to select the best-performing CF from the ensemble based on external metric evaluation. In contrast, \ourmethod’s CF generation is entirely agent-driven: for each instance, AURA intelligently explores different generation settings and editing tools (RadEdit or PRISM), autonomously produces multiple candidate counterfactuals, and evaluates them using internal metrics. It then selects the optimal counterfactual as the final output—without any external tuning or post-processing. To ensure fairness, \ourmethod\ is constrained to produce and present a maximum of five counterfactuals per instance, aligning with the ensemble conditions.

We quantitatively assess the CFs using four metrics: (i) \textit{Subject Identity Preservation} (SIP) is the L1 distance between CF and factual images to assess preservation of identity~\cite{mothilal2020explaining}; \cite{mothilal2020explaining}; (ii) \textit{Counterfactual Prediction Gain} (CPG) measures the absolute difference in an off-the-shelf DenseNet121~\cite{iandola2014densenet} multi-head classifier's predictions (from TorchXRayVision~\cite{cohen2022torchxrayvision}) between factual and CF images~\cite{nemirovsky2020countergan}. A higher CPG indicates a greater shift across the classifier’s decision boundary; 
(iii) \textit{Classifier Flip Rate} (CFR) is the number of samples with flipped predictions after CF generation~\cite{fathi2024decodex}; (iv) \textit{Structural Similarity Index Measure} (SSIM) to identify the visual similarity between factual and CF images~\cite{kumar2022counterfactual}.

Table~\ref{tab:cf_edit_metrics} presents the quantitative comparison of these methods. We observe that generating multiple counterfactuals and selecting the best (Ensemble) significantly boosts performance over single CF generation across metrics. Specifically, \ourmethod\ demonstrates a superior balance across metrics: its SSIM is notably higher than PRISM-Ensemble, indicating better identity preservation while achieving similar CPG/CFR, and its SIP is comparable to RadEdit-Ensemble. 
\begin{table}[t]
\centering
\caption{Comparison of counterfactual editing methods across evaluation metrics. Baselines (RadEdit, PRISM) include single CF generation (\#CFs=1) and an Ensemble (\#CFs=5). \ourmethod\ (\#CFs=5) acts as an intelligent agent, internally self-evaluating and selecting the optimal CF.}
\begin{tabular}{@{}lc@{\hskip 1em}c@{\hskip 1em}c@{\hskip 1em}c@{\hskip 1em}c@{}}
\toprule
\textbf{Method} & \textbf{\# CFs} &  \textbf{CPG $\uparrow$} & \textbf{CFR $\uparrow$} & \textbf{SSIM $\uparrow$} & \textbf{SIP $\downarrow$} \\ \hline
RadEdit           &      1 &  0.264     &    0.41    &      0.764  &    0.055    \\
RadEdit-Ensemble  &     5   &    0.355   &    0.55    &   0.778    &     0.059   \\
\midrule
PRISM             &     1   &   0.418    &     0.67  &    0.648 &      0.081  \\
PRISM-Ensemble    &     5   &     0.459   &     0.71   &   0.661     &    0.079    \\
\midrule
AURA              &    5    &      0.443    &     0.71   &  0.740     &       0.060    \\
\bottomrule
\label{tab:cf_edit_metrics}
\end{tabular}
\end{table}
\subsection{Adaptive Explanations with Limited Pathological Knowledge}
\begin{figure}
\centering
\includegraphics[width=\linewidth]{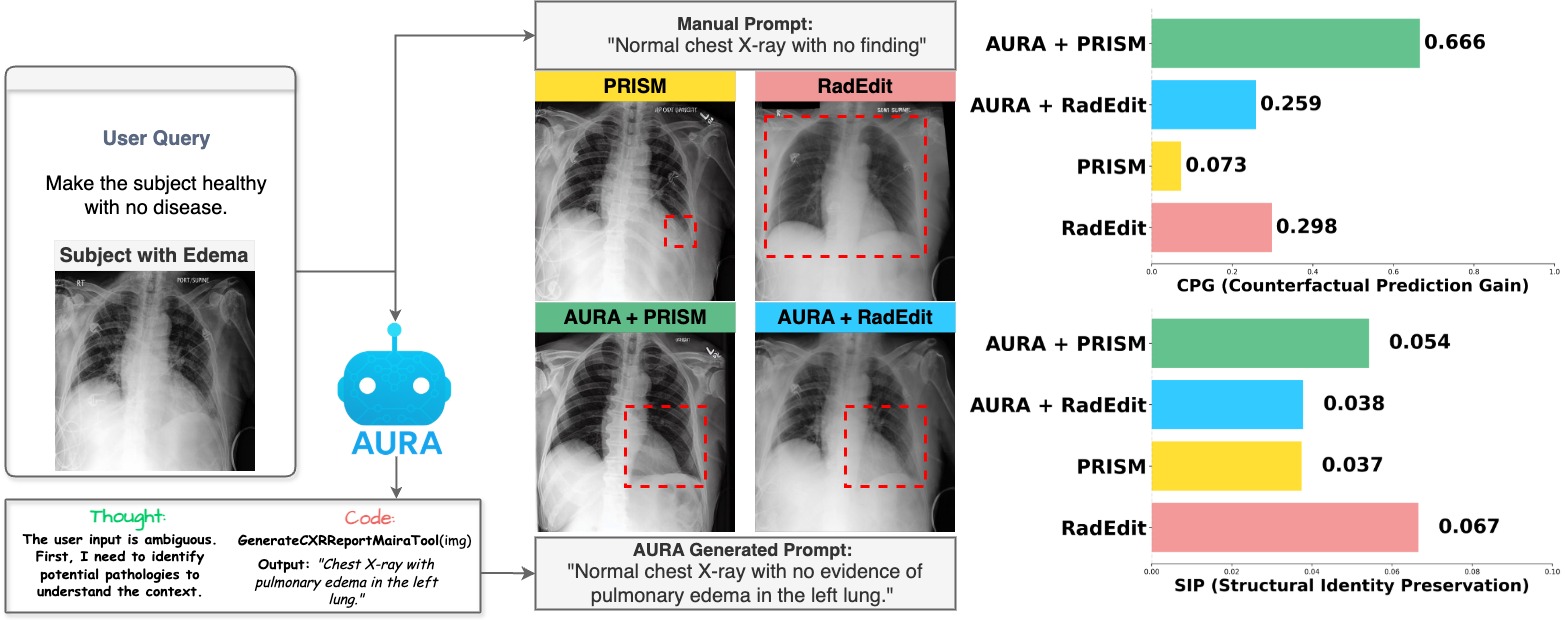}
\caption{\textbf{Adaptive CF generation from user queries.} This figure illustrates \ourmethod's agentic reasoning. Given an ambiguous query (top-left), \ourmethod\ autonomously identifies limited pathology knowledge, calls a report tool, and generates a precise prompt (bottom-middle). \textcolor{red}{Bounding boxes} indicate the area of interest for each edit. Top: Baselines (PRISM/RadEdit) utilise generic prompts, resulting in over- or under-editing respectively. Bottom: AURA-guided edits show improved focus on correct pathology and visual quality. CPG and SIP plots (right) quantitatively confirm \ourmethod's superior performance.}
\label{fig:exp2}
\end{figure}

We evaluate \ourmethod’s emergent ability to generate accurate explanations when provided with limited pathological details, a critical requirement for real-world clinical use where users' prior knowledge is limited. As illustrated in Figure~\ref{fig:exp2}, ambiguous user inputs with no explicit pathological context typically result in poor-quality edits when standard image-editing tools rely solely on generic prompts such as \textit{"Normal chest X-ray with no finding"}. These tools inherently depend on precise textual instructions to produce targeted edits. In contrast, \ourmethod\ recognizes the lack of explicit information as a challenge and proactively engages a medical report generation tool to identify specific pathological findings. It then leverages this detailed information to create an accurate prompt for subsequent editing tasks. This adaptive, agent-based reasoning enables \ourmethod\ to surpass predefined instructions, achieving edits that are both precise and contextually relevant. Quantitative results, including CPG and SIP scores in Figure~\ref{fig:exp2}, further demonstrate \ourmethod’s effectiveness in these demanding situations.

\section{Conclusion}
In this work, we introduced \ourmethod, an agentic AI system designed to analyze, explain, and evaluate chest X-ray images through a modular toolbox encompassing segmentation, counterfactual generation, and evaluation tools. Building upon recent works in LLMs, \ourmethod\ brings explainability and autonomy to medical imaging agentic workflows. Our evaluations highlight \ourmethod’s ability to operate effectively under limited supervision and incomplete pathological information, conditions that often reflect real-world clinical settings. Notably, \ourmethod\ demonstrates intelligent decision-making by recognizing gaps in context, invoking report generation tools when necessary, and refining visual edits accordingly. This capacity for self-directed reasoning and adaptation represents a step forward in the development of clinically aligned AI systems. Looking ahead, we envision agentic frameworks like \ourmethod\ playing a critical role in advancing trustworthy, transparent, and context-aware AI for medical diagnostics. 

\section{Acknowledgement}
The authors are grateful for funding provided by the Natural Sciences and Engineering Research Council of Canada, the Canadian Institute for Advanced Research (CIFAR) Artificial Intelligence Chairs program, Mila - Quebec AI Institute, Google Research, Calcul Quebec, and the Digital Research Alliance of Canada.

\clearpage
\bibliographystyle{splncs04}
\bibliography{main}
\end{document}